\documentclass[letterpaper, 10 pt, conference]{ieeeconf}  

\usepackage[colorlinks=true,linkcolor=black,urlcolor=magenta]{hyperref}
\usepackage{cite}
\usepackage{multirow}
\usepackage{booktabs}
\usepackage{pdfpages}
\IEEEoverridecommandlockouts                            
\usepackage{graphics}
\usepackage{epsfig}
\usepackage{mathptmx}
\usepackage{times}

\usepackage{amsmath}
\usepackage{amssymb} 

\title{\LARGE\bf ACRNet: Attention Cube Regression Network for Multi-view Real-time 3D Human Pose Estimation in Telemedicine}

\author{Boce Hu, Chenfei Zhu, Xupeng Ai and Sunil K. Agrawal*
\thanks{Boce Hu, Chenfei Zhu, and Xupeng Ai are with the Department of Mechanical Engineering, Columbia University, New York, NY, 10027 USA (e-mail:bh2770@columbia.edu)}
\thanks{* Sunil K. Agrawal is with the Department of Mechanical Engineering, Department of Rehabilitation and Regenerative Medicine, Columbia University, New York, NY 10027 USA (e-mail: sunil.agrawal@columbia.edu).}
\thanks{This work involved human subjects in its research. Approval of all ethical and experimental procedures was granted by the Institutional Review Board of Columbia University under Protocol No.AAAQ7781.}
}

\begin{document}

\maketitle

\begin{abstract}
Human pose estimation (HPE) for 3D skeleton reconstruction in telemedicine has long received attention. Although the development of deep learning has made HPE methods in telemedicine simpler and easier to use, addressing low accuracy and high latency remains a big challenge. In this paper, we propose a novel multi-view Attention Cube Regression Network (ACRNet), which regresses the 3D position of joints in real time by aggregating informative attention points on each cube surface. More specially, a cube whose each surface contains uniformly distributed attention points with specific coordinate values is first created to wrap the target from the main view. Then, our network regresses the 3D position of each joint by summing and averaging the coordinates of attention points on each surface after being weighted. To verify our method, we first tested ACRNet on the open-source ITOP dataset; meanwhile, we collected a new multi-view upper body movement dataset (UBM) on the trunk support trainer (TruST) to validate the capability of our model in real rehabilitation scenarios. Experimental results demonstrate the superiority of ACRNet compared with other state-of-the-art methods. We also validate the efficacy of each module in ACRNet. Furthermore, Our work analyzes the performance of ACRNet under the medical monitoring indicator. Because of the high accuracy and running speed, our model is suitable for real-time telemedicine settings. The source code is available at:\href{https://github.com/BoceHu/ACRNet}{https://github.com/BoceHu/ACRNet}
\end{abstract}

\section{INTRODUCTION}
Telemedicine is an emerging and booming treatment approach in the medical field because of its high efficiency, cost-effective strategy, and safety. Compared with traditional medical treatment, telemedicine improves treatment efficiency through timely feedback between doctors and patients. Also, it leverages technologies, such as computer-aided pose assessment, to provide accurate and objective patient conditions, during which the time of supervision and evaluation by therapists is reduced, and the number of face-to-face diagnosis sessions is also lessened, thus significantly minimizing the cost of rehabilitation. Meanwhile, telemedicine offers new probabilities for patients with reduced mobility or disabilities to be treated at home, effectively preventing infection caused by exposure to unsanitary conditions. In practice, delivering such a service remotely requires satisfying several constraints like exploiting limited computing power on personal computers, high precision, and real-time performance.

Telemedicine has been widely used in three medical application areas: prediction of movement disorders, diagnosis of movement disorders, and sports rehabilitation training \cite{9425001,wu2020human,li2020human}. One of the most significant technology for realizing them is utilizing human pose estimation (HPE) to reconstruct the 3D human body skeleton. Considering the actual implementation requirements in telemedicine, scientists proposed sensor-based and learning-based methods to estimate human pose for 3D reconstruction. However, sensor-based methods (e.g., wearable equipment) need to be attached to the body of patients, which affects patient movement, leading to inaccurate diagnoses. Moreover, appropriately adjusting devices on wearable equipment, such as inertial measurement units (IMUs) and gyroscopes, requires professional skills. Therefore, the drawbacks of sensor-based methods seriously hinder its further development in telemedicine. 

\begin{figure}
  \centering
  \centerline{\includegraphics[scale=0.65]{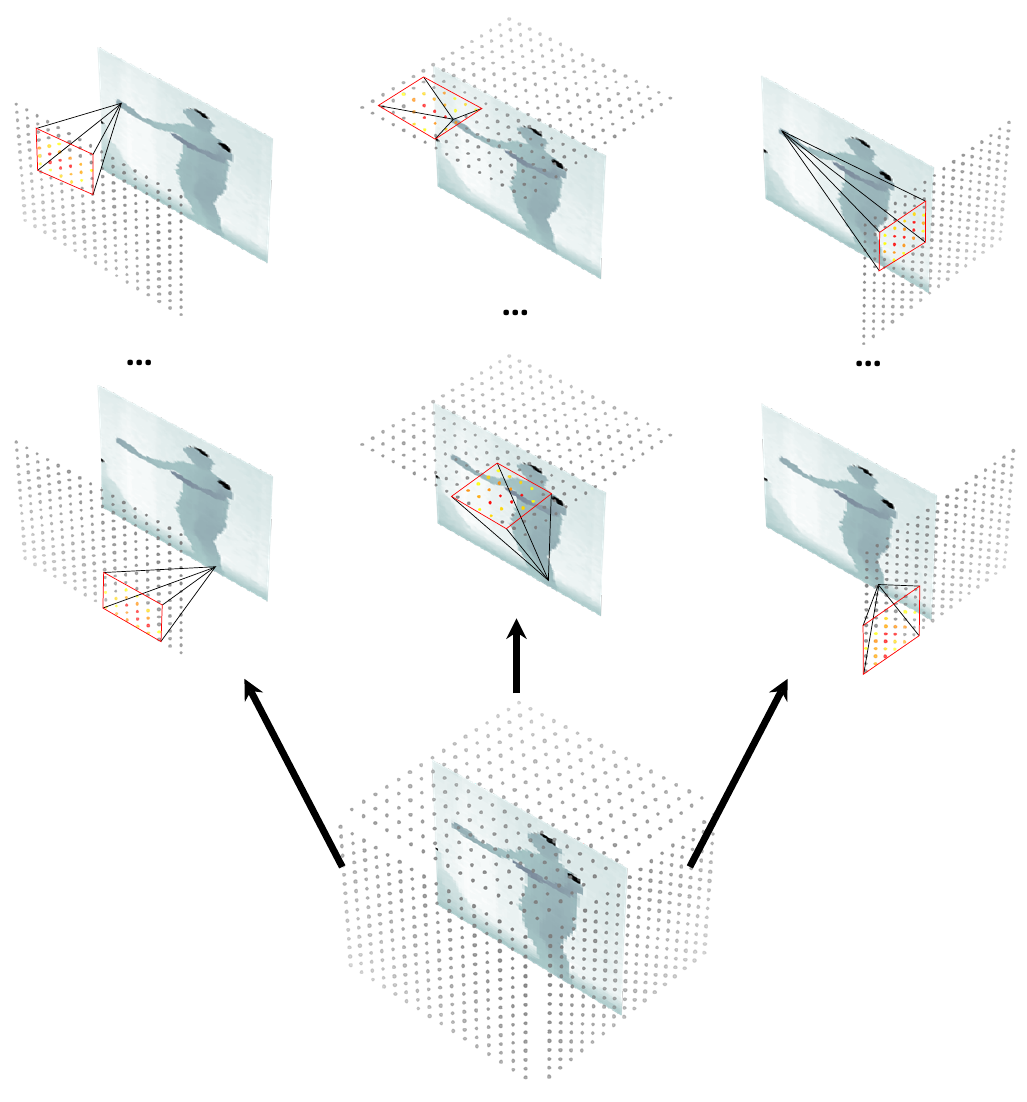}}
  \vspace{-0.2cm}
  \caption{An attention cube is introduced to wrap the target from the main view, and evenly distributed gray points stand for attention points on each surface. ACRNet calculates the point-wise weight on each surface to find the informative attention points for regressing the 3D position of joints. In the figure, the darker the point's color, the higher its weight.}
  \label{sketch}
  \vspace{-0.5cm}
\end{figure}

Benefiting from advances in deep learning and computer vision, learning-based HPE technology enables telemedicine to get rid of counting on sensor-based methods in a non-contact and easily calibrated way. Nevertheless, these methods still face low accuracy and high latency problems. As a result, to meet the multiple requirements in telemedicine, we propose a novel Attention Cube Regression Network (ACRNet), a unified and effective network with fully differentiable end-to-end training ability to perform estimation work based on multi-view depth images. ACRNet introduces an attention cube to wrap the object and aggregates information from each surface of the cube to estimate the 3D position of human joints, as shown in Fig. \ref{sketch}. More specifically, a fixed-size cube wrapping the human body from the main view will be created first, with a fixed number of points distributed uniformly on each surface. Points on the same surface constitute an attention matrix. Then, our network fuses the feature information from all views to calculate the weight matrices of all attention matrices w.r.t each joint. Finally, joints position are deduced by the sum of the element-wise products of all the attention matrices and corresponding weight matrices. Within the model, feature maps are extracted from depth images by a two phases backbone network; after that, a multi-view fusion module integrates feature maps from different views using dynamic weights according to the mechanism of cross similarity. Next, a weight distribution module simultaneously computes the attention matrix's corresponding weight matrix on each surface for the final regression. For each joint, contributions of different attention points are not equal; hence, each joint has its informative points (points with high weight) to be used to regress the position and non-informative points (points with low weight) to be discarded.

To validate our method, ACRNet is first tested on the ITOP dataset. The results demonstrate that our method outperforms the state-of-the-art methods on front-view settings while on par with the best state-of-the-art method on top-view settings. Moreover, the running speed of ACRNet achieves 92.3 FPS on a single NVIDIA Tesla V100 GPU, enabling it to work in a real-time environment. Furthermore, to verify the capability of our model in real rehabilitation scenarios, therefore providing a technical foundation for the telemedicine platform, we collect a new medical multi-view upper body movement dataset (UBM) from 16 healthy subjects on the trunk support trainer (TruST) \cite{7872386}, labeled by a Vicon infrared system. Our model consistently outperforms the baseline \cite{Xiao_2018_ECCV} on this dataset.
Overall, the contributions of this manuscript are:
\begin{itemize}
    \item ACRNet: A fully differentiable multi-view regression network based on depth images to estimate 3D human joint positions for telemedicine use.
    \item A new backbone structure and a dynamic multi-view fusion module are proposed. Both of them improve the representation ability of our model.
    \item UBM: A Vicon-labeled multi-view upper body movement dataset for rehabilitation use, consisting of depth images collected from 16 healthy subjects.
\end{itemize}

\section{RELATED WORKS}
\label{sec:related}
\subsection{3D HPE with Sensor-based Methods}
Currently, clinical diagnosis and treatments using motion capture and pose estimation depend on Vicon because of its preciseness, but this system is unsuitable for telemedicine caused of its expensive components and difficulty transferring. Thus, Sensor-based wearable equipment is used in telemedicine to capture patients' motion data. Li \textit{et al.}\cite{9099531} use multiple inertial sensors attached to the lower limbs of children with cerebral palsy to evaluate their motor abilities and validate therapy effectiveness. Sarker \textit{et al.}\cite{s22062300} infer the complete upper body kinematics for rehabilitation applications based on three standalone IMUs mounted on wrists and pelvis. Nguyen \textit{et al.}\cite{5409556} propose using optical linear encoders and accelerometers to capture the goniometric data of limb joints. As these methods will affect patients' movement, and some components are also hard to calibrate, they will lead to an inaccurate diagnosis, weakening its application value in telemedicine.

\subsection{3D HPE with Learning-based Methods}
Learning-based HPE methods can be divided into machine learning and deep learning. The former \cite{5995316,Rafi_2015_CVPR_Workshops,10.1145/2398356.2398381,Jung_2015_CVPR} usually transforms the estimation problem into a classification problem by calculating the probability of the location for each joint. A serious drawback of these methods is the severely deficient representation ability when the estimation work is complex. As a result, deep learning methods utilizing RGB or depth images have become mainstream in this field. RGB-images-based methods\cite{8765346,He_2017_ICCV,Xiao_2018_ECCV,Yuan_2021_CVPR} are intuitive and convenient. Nevertheless, the accuracy of those methods is relatively low due to the lack of spatial information. With the popularity of depth cameras, depth-image-based methods address this shortcoming. Guo \textit{et al.}\cite{WANG2018404} propose a tree-structured Region Ensemble Network to aggregate the depth information. Kim \textit{et al.}\cite{KIM2020107462} estimate human pose by projecting the depth and ridge data in various directions. Qiu \textit{et al.}\cite{qiu2019cross} tackle the core problems of monocular HPE, like self-occlusion and joint ambiguity, by an embedded fusion layer that merges features from different views. He \textit{et al.}\cite{he2020epipolar} extend this method with the Transformer to match the given view with neighboring views along the epipolar line by calculating feature similarity to obtain the final 3D features. Further, Moon \textit{et al.}\cite{8578631} and Zhou \textit{et al.}\cite{9107101} take advantage of point clouds with more intuitive information transformed from depth images to acquire an exact 3D position of the human body. Although point-cloud-based methods are accurate enough, these methods generate a plethora of parameters during execution, consuming more time and memory, which prevents them from working in real-time. Consequently, considering the pros and cons of different data types, our work will directly adopt the depth map as the model's input.

Inspired by the work \cite{xiong2019a2j}, which exploits the global-local spatial information from 2D anchor points, we introduce the 3D attention cube. Our attention points are created following their method; however, we enhance the correlation between 3 principle directions by facilitating the interaction of different cube surfaces to eliminate the estimated bias of each surface. This mutually constrained property improves the robustness of our method.
\begin{figure*}[!htb]
  \centering
    \centerline{\includegraphics[scale=0.58]{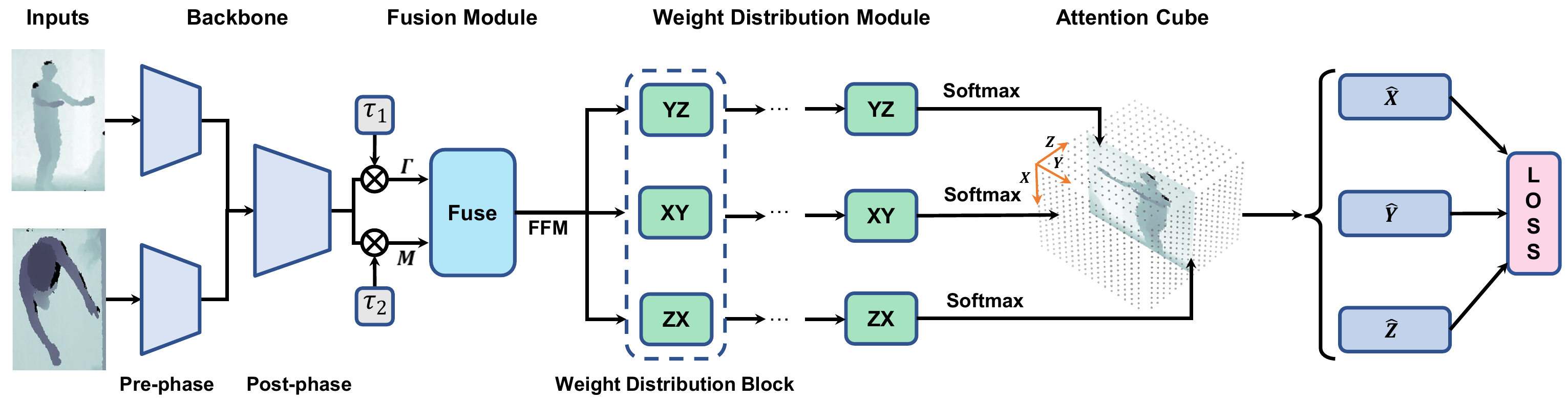}}
    \vspace{-0.1cm}
    \caption{Overview of the workflow of the ACRNet. ACRNet includes an embedded multi-view feature fusion module driven by a feature extraction backbone and a weight distribution module with an attention cube at the end. $\tau_1$ and $\tau_2$ are weighting factors to differentiate the main view $\Gamma$ and the auxiliary view $M$. $FFM$ denotes fused feature maps. Each weight distribution block contains 3 identical Transformers for different surfaces.}
    \vspace{-0.35cm}
    \label{pipeline}
\end{figure*}

\section{METHODOLOGY}
\label{sec:method}
The workflow of our ACRNet is shown in Fig. \ref{pipeline}. Given images captured by two depth cameras simultaneously, ACRNet first extracts the feature map of each view by the backbone network and then merges feature maps from two views as the fused feature maps (FFM) using the fusion module. After that, the weight distribution module calculates the weight of each attention point on three surfaces respectively. The 3D joint position is estimated by summing and averaging the element-wise products of attention matrices and corresponding weight matrices on each surface. In the following part, we will elaborate on the modules of ACRNet.

\subsection{ACR: Attention Cube Regression Network}
\label{ACRNetwork}
\textbf{Backbone Network:}
Following Swin-Transformer\cite{liu2021swin} and Convnext\cite{https://doi.org/10.48550/arxiv.2201.03545}, we adjust three aspects of the vanilla SE-ResNet-50\cite{hu2018squeeze} as our backbone:
   (a) the number of blocks in this four-stage network is rearranged from (3, 4, 6, 3) to (3, 3, 9, 3), and the stride is set to 1 in the last stage to reserve more information.
   (b) the Rectified Linear Unit (ReLU)\cite{nair2010rectified} activation is altered with the Gaussian Error Linear Units (GELU)\cite{https://doi.org/10.48550/arxiv.1606.08415} activation, which reserve gradients while bounding the effect in the negative domain to stop neurons from inactivating to allow for better mixing of features between layers.
   (c) the number of activation functions in each block is reduced by only keeping the activation function between the second and the third layer in each block.

In addition to the layer-wise modification above, we regather the four-stage backbone into two phases: the pre-phase block and the post-phase block. The former consists of the first two stages, and the latter includes the rest. Because the information about objects differs by the view, we utilize different pre-phase blocks to extract features from different perspectives. Feature maps from all views are then fed into the same post-phase block to extract higher dimensional features. Compared with the original SE-ResNet-50, our backbone performs better without adding mass parameters.

\begin{figure}[t]
    \centering
    \includegraphics[scale=0.55]{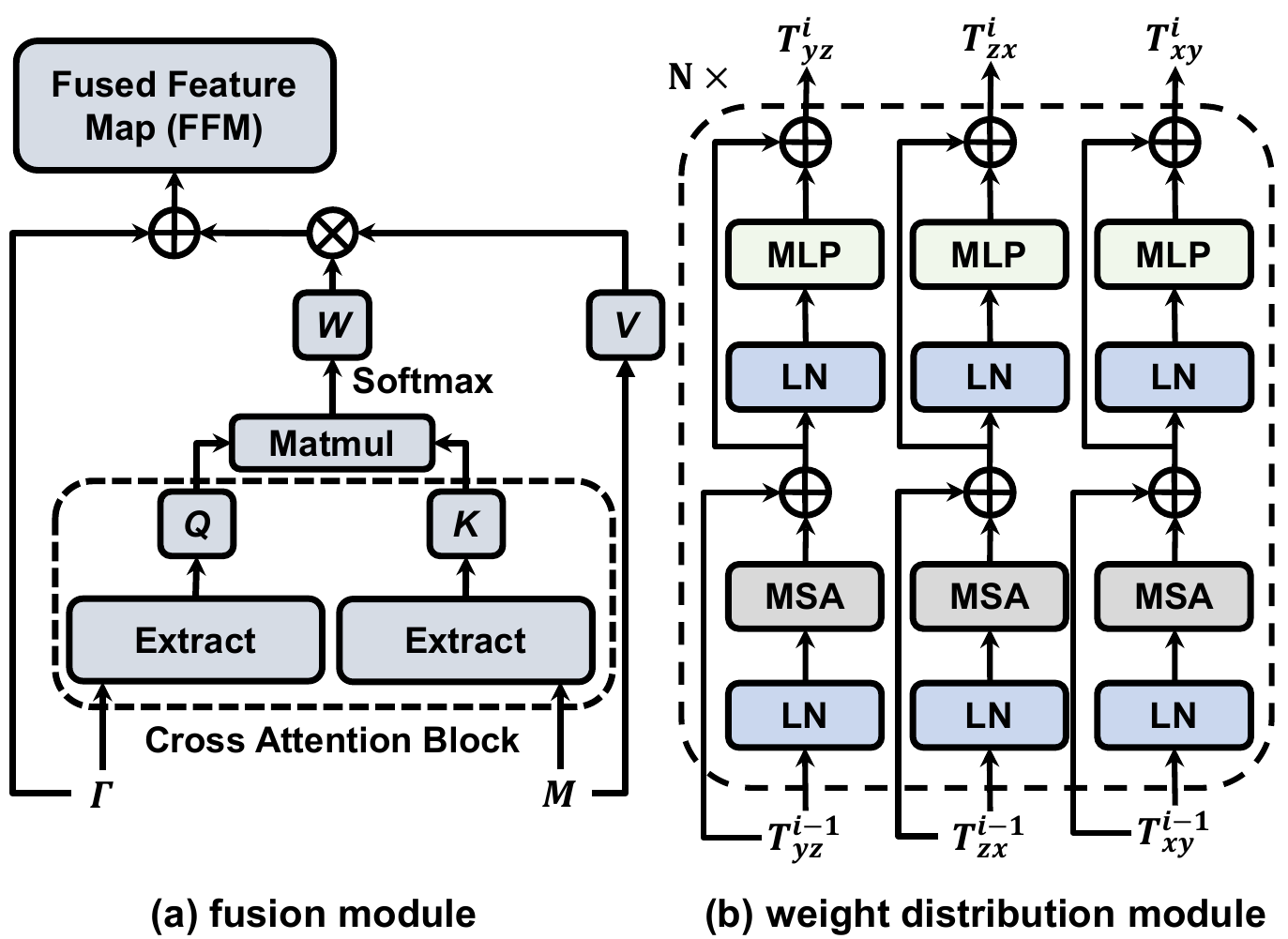}
    \vspace{-0.1cm}
    \caption{(a) Given feature maps from the main view $\Gamma$ and the auxiliary view $M$, the cross attention block computes the cross attention to dynamically fuse feature maps according to the correlation between $\Gamma$ and $M$. "$\oplus$" denotes the element-wise sum, "$\otimes$" denotes element-wise multiplication, and "$Matmul$" represents matrix multiplication. (b) Three branches enclosed by the dotted line consist of one weight distribution block. Each branch follows the architecture of the vanilla Transformers.}
    \label{fig:fusion}
    \vspace{-0.3cm}
\end{figure}

\textbf{Fusion Module:}
This module aims to merge features from different views, as shown in Fig.\ref{fig:fusion}(a). To illustrate the pipeline, we use $\Gamma=\{\gamma_1,\gamma_2,...,\gamma_n\}$ and $ M=\{\mu_1,\mu_2,...,\mu_n\}$ to denote feature maps generated by the backbone from the main and auxiliary views. Here, $\gamma_i$ and $\mu_i$ indicate the corresponding feature map of $i_{th}$ channel. Next, $\Gamma$ and $M$ are fed into the cross-attention block, in which we use a novel method to dynamically extract information from the auxiliary view to complement the main view. More specifically, we compute one-to-one matching query $q_i$ extracting from $\gamma_i$ and key $k_i$ extracting from $\mu_i$ on each channel to determine their cross-similarity, as shown below:
\begin{align}
    Q &= \{q_1,q_2,...,q_n\} \text{ where } q_i = Extract(\gamma_i)\\
    K &= \{k_1,k_2,...,k_n\} \text{ where } k_i = 
    Extract(\mu_i) \\
    W &= \{\omega_1,\omega_2,...,\omega_n\} = Softmax(\frac{Q K^T}{\sqrt{d}})
\end{align}
where \textit{Extract}$(\cdot)$ is three convolution layers to filter feature maps further, \textit{$\sqrt{d}$} is a scaling factor according to the number of feature channels, and $W$ represents the cross-similarity of all channels. In this way, the fusion module calculates the weight matrix $\omega_i$ considering the correlation between each $\gamma_i$ and $\mu_i$ for dynamic merging. After that, the value $V$, equal to $M$, will be element-wise multiplied by $W$ as the final feature maps obtained from the auxiliary view. Accordingly, the fused feature maps (\textit{$FFM$}) can be achieved by:
\begin{align}
FFM = \Gamma + W \odot V
\end{align}

The previous work\cite{qiu2019cross} inspires our embedded fusion module, but we improve the merge procedure with the cross-attention mechanism, which leverages the interdependence between different views more appropriately. Because of the end-to-end training attribute, our fusion module can be inserted into any location on other networks.

\textbf{Weight Distribution Module:}
The weight distribution module comprises \textit{N} weight distribution blocks, each block contains 3 identical branches for different surfaces: $XY, YZ, ZX$, as shown in Fig. \ref{fig:fusion}(b). In our implementation, $\textit{N}$ is set to 4. Each branch follows conventional Transformers, which can be formulated as follows:
\begin{align}
T^{i'} &=T^{i-1}+MSA(LN(T^{i-1}))\nonumber, \\
T^{i} &=T^{i'}+MLP(LN(T^{i'}))
\end{align}
where $MSA(\cdot)$ denotes the Multi-head Self-Attention, $MLP(\cdot)$ denotes Multi-Layer Perceptron, $LN(\cdot)$ denotes Layer Normalization, $T^{i-1}$ represents the input features of the current block, as well as the output features of the previous block and $T^{i}$ stands for the output features of the current block. The input size of this module is determined by the fusion module, and the output size equals the attention matrix size. Therefore, after multi-block computation, the feature map output by the fusion module will eventually be transformed into the weight matrix of its corresponding attention matrix, which means each attention point has its own weight to determine its importance.

\textbf{Attention Cube Regression:}
We use a cube with evenly distributed attention points on each surface to wrap the main view, as shown in Fig. \ref{sketch}. In our implementation, the number of attention points on each surface is set to $64\times64=4096$. For each attention point on a cube surface, we only consider its in-plane coordinate and ignore the coordinate along the normal direction. For example, points on the surface orthogonal to the z-axis only have x-axis and y-axis coordinates. Since each joint will be projected on three surfaces: $XY, YZ, ZX$, our model aims to find the informative attention points on each surface to regress the joint position. To be more specific, the 3D position of joint $\textit{j}$ can be attained by:
\begin{equation}
\left\{
\begin{aligned}
x_j &=Mean({\sum_{x\in {XY}}\eta_{j}^{x}}{P_{j}^{x}}+{\sum_{x\in {ZX}}\eta_{j}^{x}}{P_{j}^{x}}) \\
y_j &=Mean({\sum_{y\in {XY}}\eta_{j}^{y}}{P_{j}^{y}}+{\sum_{y\in {YZ}}\eta_{j}^{y}}{P_{j}^{y}}) \\
z_j &=Mean({\sum_{z\in {YZ}}\eta_{j}^{z}}{P_{j}^{z}}+{\sum_{z\in {ZX}}\eta_{j}^{z}}{P_{j}^{z}})
\end{aligned}
\right.
\end{equation}
where the subscript of the summation symbol indicates the surface to be calculated on; $\eta_{j}^{x}$, $\eta_{j}^{y}$, and $\eta_{j}^{z}$ are the weight matrices of joint $j$ on their corresponding calculated surfaces, output by the weight distribution module and normalized by the softmax function; ${P^{x}}$, ${P^{y}}$, and ${P^{z}}$ denote the x-value, y-value, and z-value of all attention points on that surface. After summing the dot products of weight matrices and corresponding attention matrices from two surfaces, the average is calculated to reduce the bias.

The designed attention cube has excellent flexibility since it can regress joint positions in 2D pixel coordinates and 3D external coordinates, and the difference between the two circumstances is only how the cube is created. In detail, the first way applies pixel coordinates on depth images to regress the x and y values and the world coordinate to regress the z value directly. Then, the x and y values are transformed into world coordinates according to the z value and the camera's internal parameters. To make the magnitude of values in the world coordinates comparable to pixel coordinates, we obey the work of \cite{xiong2019a2j} to multiply the depth value by a parameter. For this regression way, the side lengths of the x-axis and y-axis directions of the cube, as shown in Fig. \ref{pipeline}, are set equal to the height and width of the depth map. The side length of the z-axis direction is dataset specific, which covers the entire working space of the target in the depth direction under the world coordinate. However, for some medical situations, specific joints may not always appear in 2D depth images, even in the synchronized multi-view setting. In this case, it is hard to label them correctly in pixel coordinates, making it impossible for the model to predict the 3D position effectively, possibly affecting medical test results. So the second way is designed to purely apply the world coordinates to regress the 3D positions of joints. Thus, all side lengths of the cube are set to cover the entire working area of the target in the corresponding direction of that fixed external coordinate system.
\subsection{Regression Loss Design}
Following the previous work \cite{WANG2018404}, we adopt the smooth $L_1$ loss for each axis:
\begin{align}
smooth_{L1}(x) =
\begin{cases}
0.5x^2  & \text{if $\lvert x \rvert <$ $\beta$} \\
x-0.5 & \text{otherwise}
\end{cases}
\end{align}
where x is the distance between the ground truth and the predicted position along the specific axis. $\beta$ is the threshold set to 1 in the pixel coordinate and 3 in the word coordinate. Consequently, the end-to-end training loss function for ACRNet is:
\begin{align}
\label{loss}
Loss = loss_{x}+loss_{y}+\lambda loss_{z}
\end{align}
where \textit{Loss} is the aggregated loss; items at the right of Eqn. \ref{loss} is the smooth $L_1$ loss of each axis, and $\lambda$ is set to 3 to mitigate the noise of depth value.

\section{EXPERIMENTS}
This section first introduces datasets and evaluation metrics and provides training and testing details. After that, comparisons between ACRNet and state-of-the-art methods are drawn to verify the superiority of our methods. We also conduct exploration studies to explore the effectiveness of the network architecture and finally analyze the performance of ACRNet under the medical monitoring indicator.   
\subsection{Datasets}
\textbf{ITOP Multi-viewed Dataset \cite{haque2016towards}:} This dataset contains 40K training images and 10K test images of the front and top views of the human body collected from 20 subjects. Two Asus Xtion PRO cameras capture images simultaneously to ensure the synchronization of each frame. ITOP dataset focuses on a series of identical body actions, and each depth image is annotated with 15 body joints.

\begin{figure}[t]
    \centering
    \includegraphics[scale=0.63]{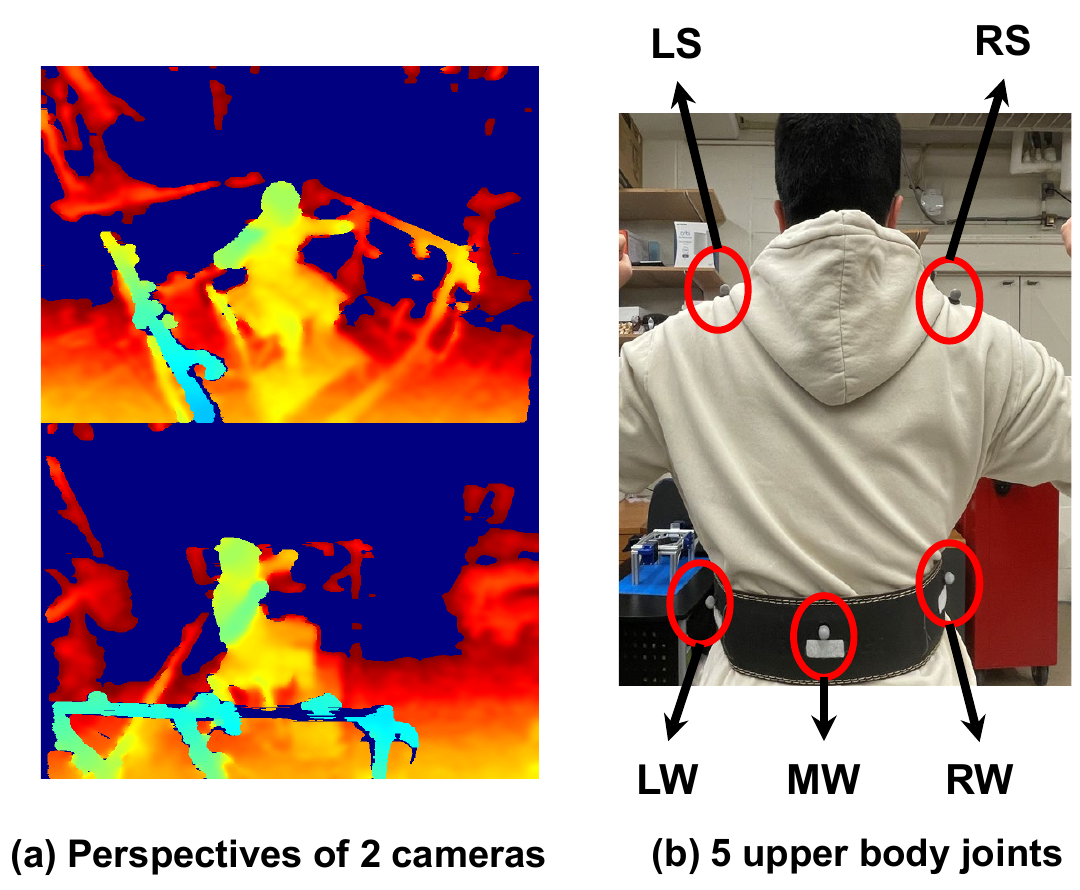}
    \vspace{-0.1cm}
    \caption{(a) Colorized depth images from the right-front view (above) and the right view (below). (b) 5 upper body joints marked with Vicon markers are the left waist (LW), mid waist (MW), right waist (RW), left shoulder (LS), and right shoulder (RS). The first three markers are stuck on the belt, which is placed on the participants' lower ribs (thoracic region: T9-12).}
    \label{fig:UBM}
    \vspace{-0.5cm}
\end{figure}

\textbf{Our UBM Multi-viewed Dataset:} To imitate the real rehabilitation scenario, we collected this dataset with the assistance of TruST \cite{7872386}, a cable-driven robotic device for rehabilitation, training, and strengthening of the upper body. The concept behind this dataset is the Star Excursion Balance Test (SEBT), a widely used postural task for objectively measuring the upper body postural limits of patients with lower extremity injuries \cite{gribble2012using}. UBM consists of 156K trunk movement depth images collected from 16 subjects by two Intel RealSense D435i depth cameras placed at the right and right-front sides of the subject. The colorized depth images from two perspectives are shown in Fig. \ref{fig:UBM}(a). Depth images of 12 subjects form the training set; others constitute the test set. During data collection, all subjects sit on the bench and are instructed to perform 12 rounds of the same random sequence of upper body movements in 8 principal directions: front, rear, left, right, left-front, right-front, left-rear, and right-rear. The subject moved the trunk as far as possible in the instructed direction, returned to the neutral position, and waited for instructions to move in the next direction. The Vicon infrared motion capture system provides the external coordinates of 5 upper body joints, as shown in Fig. \ref{fig:UBM}(b), by infrared markers on each joint as labels in world coordinates. Setting cameras in these positions aims to create an information-missing environment to simulate that cameras are not being mounted at the optimal position as expected. Also, due to the complex architecture of TruST, the experiment environment is relatively noisy, resulting in an unclean camera background, similar to the actual situation of in-home therapy utilizing some paramedical equipment. Altogether, all settings designed in our experiments are to make our experiment environment as equivalent to the actual situation of telemedicine treatment as possible.

\begin{table*}[!t]
\setlength\tabcolsep{3.5pt}
\normalsize
\centering
\caption{Performance comparison between ACRNet and state-of-the-art methods for ITOP front-view and top-view (metric: 10cm mAP)}
\label{tab:itop}
\scalebox{0.62}{
\begin{tabular}{cccccccccc|cccccccccc}
\toprule
\multicolumn{10}{c|}{ITOP front-view} & \multicolumn{9}{c}{ITOP top-view}\\
\hline
\multirow{2}*{Body part}    &RF     &RTW    &IEF    &VI     &REN    &A2J    &V2V    &DECA-D3    &ACR    &RF &RTW    &IEF    &VI     &REN    &A2J    &V2V    &DECA-D3    &ACR\\
~&\cite{5995316}&\cite{Jung_2015_CVPR}&\cite{7780881}&\cite{haque2016towards}&9$\times$6$\times$6 \cite{WANG2018404}&\cite{xiong2019a2j}&\cite{8578631}&\cite{garau2021deca}&(ours)&\cite{5995316}&\cite{Jung_2015_CVPR}&\cite{7780881}&\cite{haque2016towards}&9$\times$6$\times$6 \cite{WANG2018404}&\cite{xiong2019a2j}&\cite{8578631}&\cite{garau2021deca}&(ours)\\
\hline
Head        &63.80  &97.80  &96.20  &98.10  &98.70  &98.54  &98.29  &93.87  &98.52
            &95.40  &98.40  &83.80  &98.10  &98.20  &98.38  &98.40  &95.37  &97.82\\
Neck        &86.40  &95.80  &85.20  &97.50  &99.40  &99.20  &99.07  &97.90  &98.89
            &98.50  &82.20  &50.00  &97.60  &98.90  &98.91  &98.91  &98.68  &98.60\\
Shoulders   &83.30  &94.10  &77.20  &96.50  &96.10  &96.23  &97.18  &95.22  &96.96
            &89.00  &91.80  &67.30  &96.10  &96.60  &96.26  &96.87  &96.57  &95.82\\
Elbows      &73.20  &77.90  &45.40  &73.30  &74.70  &78.92  &80.42  &84.53  &84.14
            &57.40  &80.10  &40.20  &86.20  &74.40  &75.88  &79.16  &84.07  &81.92\\
Hands       &51.30  &70.50  &30.90  &68.70  &55.20  &68.35  &67.26  &56.49  &68.90
            &49.10  &76.90  &39.00  &85.50  &50.70  &59.35  &62.44  &54.33  &66.73\\
Torso       &65.00  &93.80  &84.70  &85.60  &98.70  &98.52  &98.73  &99.04  &98.56
            &80.50  &68.20  &30.50  &72.90  &98.10  &97.82  &97.78  &99.46  &98.01\\
Hip         &50.80  &90.30  &83.50  &72.00  &91.80  &90.85  &93.23  &97.42  &92.69
            &20.00  &55.70  &38.90  &61.20  &85.50  &86.88  &86.91  &97.42  &90.07\\
Knees       &65.70  &68.80  &81.80  &69.00  &89.00  &90.75  &91.80  &94.56  &90.34
            &2.60   &53.90  &54.00  &51.60  &70.00  &79.66  &83.28  &90.84  &82.32\\
Feet        &61.30  &68.40  &80.90  &60.80  &81.10  &86.91  &87.60  &92.04  &85.55
            &0.00   &28.70  &62.40  &51.50  &41.60  &58.34  &69.62  &81.88   &76.25\\
Upper Body  &-      &-      &-      &84.00  &-      &-      &-      &83.03  &88.44
            &-      &-      &-      &91.40  &-      &-      &-      &83.00  &87.04\\
Lower Body  &-      &-      &-      &67.30  &-      &-      &-      &95.30  &89.53
            &-      &-      &-      &54.70  &-      &-      &-      &91.39  &82.87\\
\hline
Mean        &65.80  &80.50  &71.00  &77.40  &84.90  &88.00  &88.74  &88.75  &\textbf{88.87}
            &47.40  &68.20  &51.20  &75.50  &75.50  &80.50  &83.44  &\textbf{86.92}  &85.38\\
\bottomrule
\end{tabular}}
\end{table*}

\begin{figure*}[!htp]
    \centerline{\includegraphics[width=0.8\textwidth,height=0.2\textwidth]{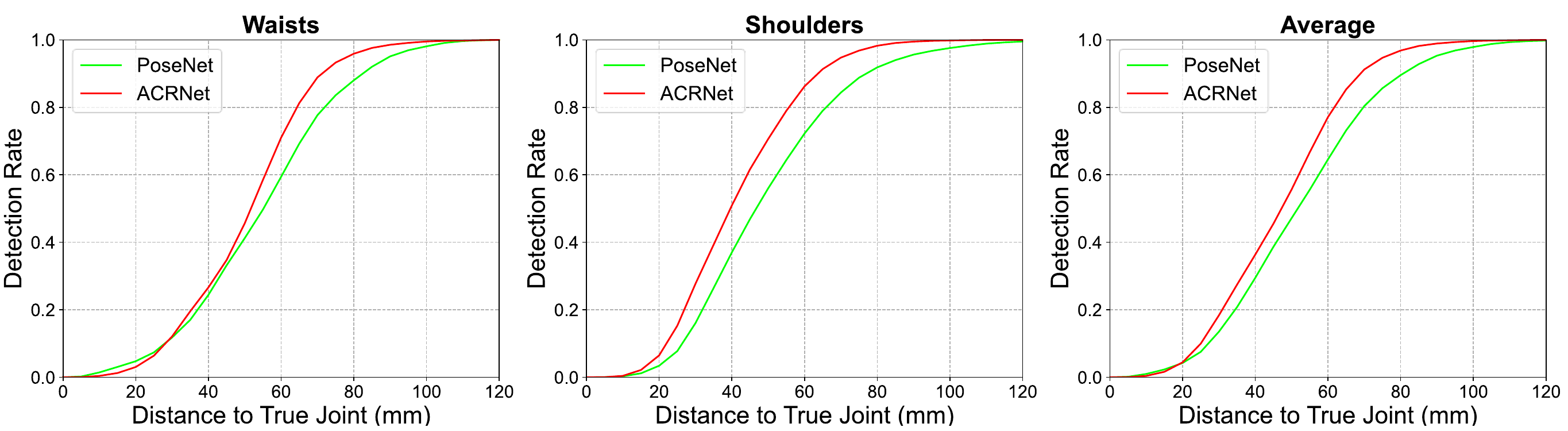}}
    \vspace{-0.2cm}
  \caption{Percentage of detected joints (PDJ) on UBM dataset for waists, shoulders, and the average. We compare the ACRNet with the baseline-PoseNet50.}
  \label{pdj}
  \vspace{-0.2cm}
\end{figure*}

\begin{table}[htp]
\setlength\tabcolsep{2.5pt}
\centering
\caption{Quantitative comparison with the baseline \cite{Xiao_2018_ECCV} on the UBM. The metric is MPJPE in millimeters}
\label{tab:mpjpe}
\scalebox{0.9}{
\begin{tabular}{c|ccccc|c}
\toprule
MPJPE   &RW  &LW  &MW   &RS  &LS    &Avg\\
\hline
Baseline-PoseNet50     &61.73  &48.92   &53.92   &43.29   &56.27    &52.83\\
ACRNet (Ours)         &53.42  &48.68   &50.49   &43.50   &39.63    &\textbf{47.15}\\
\bottomrule
\end{tabular}}
\vspace{-0.15cm}
\end{table}
\subsection{Evaluation Metrics}
Based on previous works, we use 3 metrics to evaluate our model. For the ITOP, we choose the mean average precision (mAP) to assess the success rate of detecting joints. When the Euclidean distance between the predicted joint and the ground truth is less than 10 cm, this prediction is considered successful. For the UBM, the evaluation work is carried out using the Mean Per Joint Position Error (MPJPE) and the Percent of Detected Joints (PDJ).

\subsection{Implementation Details}
A series of pre-processing methods are applied before training. First, we crop depth images to extract the human body in ITOP, following the work of Moon \cite{8578631}. For UBM, we crop the images by a fixed bounding box covering the entire working space. Second, we use a depth threshold to remove the background to segment the target. Then, images are resized to a resolution of $256 \times 256$ pixels after normalizing the depth value of each pixel to [-1,1]. We use the first and second ways mentioned above to regress ITOP and UBM, respectively. No data augmentation method is used in our implementation. We use Adam as the optimizer with a mini-batch size of 8. The initial learning rate is set to $1e^{-4}$, which decays by 0.1 every 10 epochs. The model is trained from scratch for 45 epochs on the ITOP dataset and 10 on the UBM. Our model is implemented using the Tensorflow2 framework, trained and tested on a single server.

\subsection{Comparison with State-of-the-art Methods}
\textbf{Comparisons on ITOP Dataset:} We compare ACRNet against common state-of-the-art methods based on single frame depth images\cite{5995316,Jung_2015_CVPR,7780881,haque2016towards,WANG2018404,xiong2019a2j,8578631,garau2021deca}, as in Table \ref{tab:itop}. We regard labels from the main view as ground truth in our method. The results show that ACRNet is superior to all existing methods for the front view by a large margin. For the top view, our method is on par with the DECA-D3\cite{garau2021deca} and outperforms other methods by at least 1.94 mAP. Since DECA adopts a fast Variational Bayes capsule routing algorithm to achieve viewpoint invariance, which can be considered prior knowledge, this method is more suitable for the invisible lower body in the top view; however, for the visible upper body in the top view, the performance of ours still surpasses it. In general, the representation ability of our ACRNet is remarkable, and we prove that the efficiency of ACRNet can satisfy the real-time working environment in section \ref{exp2}.

\begin{figure}[t]
  \centering
    \centerline{\includegraphics[scale=0.31]{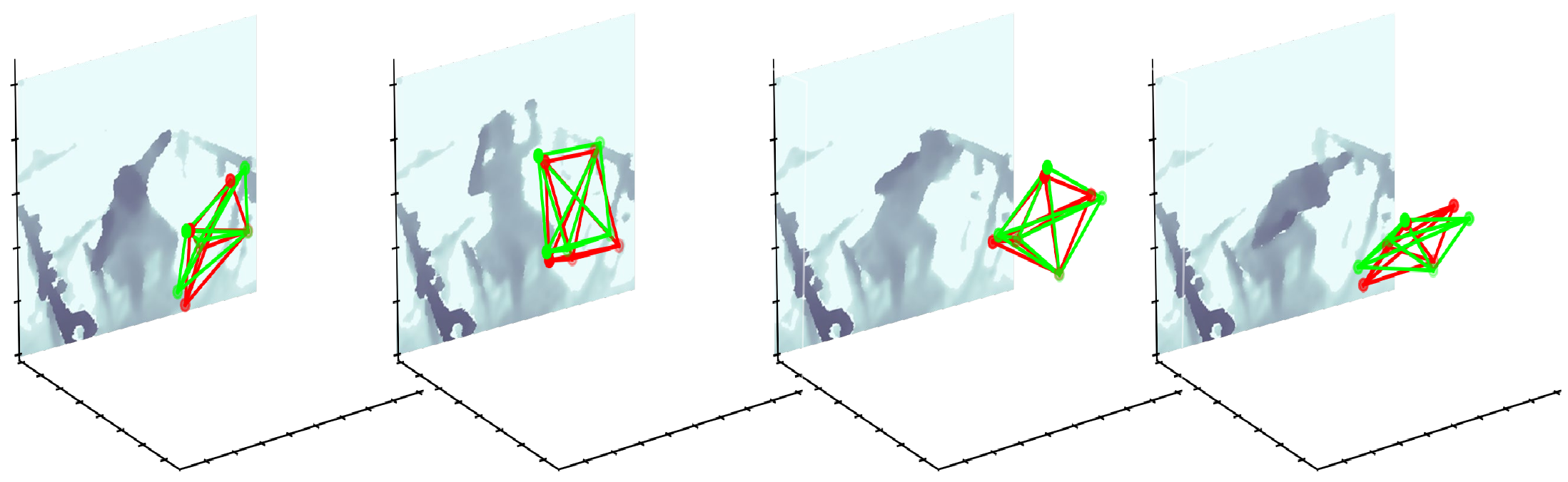}}
    \vspace{-0.3cm}
  \caption{Qualitative results on UBM. Ground truth is shown in red, and the prediction is in green. Our model still struggles to predict pose on extremely severe occlusion during specific movements (right image).}
  \label{ubmresults}

\end{figure}

\textbf{Comparisons on Our UBM Dataset:}
We compare ACRNet with the baseline \cite{Xiao_2018_ECCV}, a widely used network for evaluating new models in HPE, on UBM. The MPJPE comparison is listed in Table \ref{tab:mpjpe}, and the PDJ performance is shown in Fig. \ref{pdj}. The results can be summarized as ACRNet outperforms the baseline on both MPJPE and PDJ, reflecting that our model has a minor prediction error and is more stable. Since this dataset uses the second way for regression, it proves that creating the cube under external coordinates and directly regressing the 3d coordinates is reasonable and valid. Fig. \ref{ubmresults} shows some success and failure cases.

\subsection{Exploration Study}
\label{exp2}
To explore the effectiveness and generality of ACRNet, we conducted 2 experiments on the ITOP dataset because it is an open-source dataset and is a widely used benchmark to prove the ability of methods in HPE. 
\begin{table}[tp]
\setlength\tabcolsep{20pt}
\large
\centering
\caption{Module effectiveness analysis within ACRNet}
\label{tab:strategy}
\scalebox{0.65}{
\begin{tabular}{c|c}
\toprule
Strategy                    &mAP\\
\hline
SE-ResNet                                               &82.24\\
+Attention Cube                                         &87.04\\
+Smooth L1 Loss                                         &87.12\\
+Modified Backbone                                      &87.56\\
+Multi-view Fusion Module                               &88.35\\
+Weight Distribution Module                             &\textbf{88.87}\\
\bottomrule
\end{tabular}}
\vspace{-0.25cm}
\end{table}

\begin{figure}[t]
  \centering
    \centerline{\includegraphics[scale=0.26]{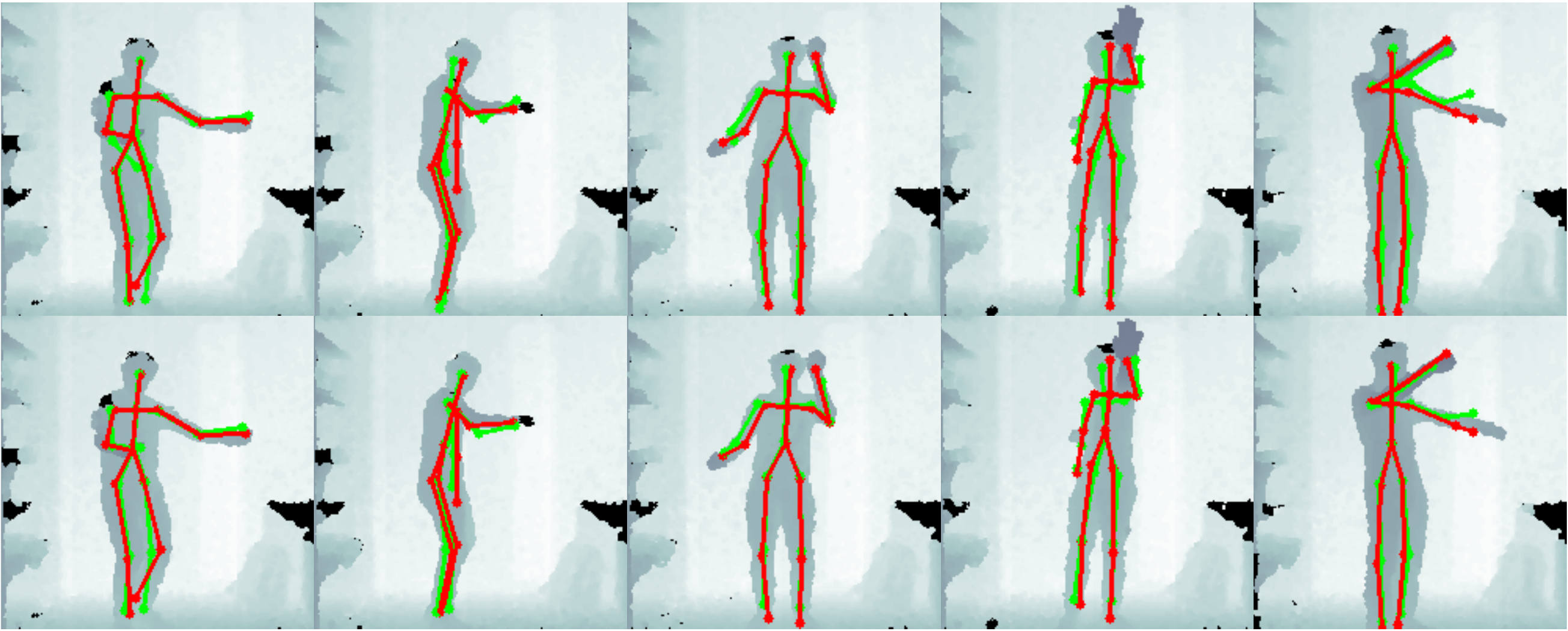}}
    \vspace{-0.2cm}
  \caption{Qualitative results comparison between the primary network (the first row, corresponding to the first row in Table. \ref{tab:strategy}) and ACRNet (the second row, corresponding to the sixth row in Table. \ref{tab:strategy}) on ITOP. Ground truth is shown in red, and the predicted pose is green.}
  \vspace{-0.4cm}
  \label{comparison}
\end{figure}

\textbf{Investigation of Module Effectiveness:}
In section \ref{sec:method}, the role of different modules is introduced, but the validity of each module still needs to be quantitatively tested. In this experiment, we incrementally add five modules on vanilla SE-ResNet-50 to finally form our ACRNet. The performance of intermediate models equipped with added modules are listed in Table \ref{tab:strategy}. Qualitative comparisons between ACRNet and the primary network (SE-ResNet-50) are shown in Fig. \ref{comparison}. The result shows that: 
\begin{itemize}
    \item By implementing our attention cube module, mAP is significantly improved from 82.24 to 86.82. It proves that our attention cube can extract more comprehensive feature information of different joints from the same backbone compared to the original dense layer in basic SE-ResNet-50. 
    \item Our modified backbone and smooth L1 loss are more suitable for depth-image-based human pose estimation work. Meanwhile, the multi-view fusion module can robustly obtain more relevant feature information from different views. The last three weight distribution modules take advantage of the attention mechanism to calculate the weight matrix focusing on specific areas of feature maps, thus guiding the reasoning process.
\end{itemize}
\begin{table}[tp]
\setlength\tabcolsep{4pt}
\large
\centering
\caption{Performance comparison of backbone networks}
\label{tab:backbone}
\scalebox{0.65}{
\begin{tabular}{c|c|c|c|c}
\toprule
Architecture        &Map Size       &Params(FLOPs)  &FPS    &mAP\\
\hline
ResNet-18           &8$\times$8     &34M (11.96G)   &306.20      &86.67\\
ResNet-34           &8$\times$8     &44M (21.63G)   &257.34      &86.70\\
ResNet-50           &8$\times$8     &63M (36.83G)   &149.13      &86.90\\
SE-ResNet-50        &8$\times$8     &66M (36.87G)   &132.43      &87.47\\
\hline
SE-ResNet-A        &16$\times$16   &66M (64.22G)   &101.58      &88.48\\
SE-ResNet-B         &16$\times$16   &72M (68.70G)   &98.01      &88.53\\
SE-ResNet-C         &16$\times$16   &72M (68.78G)   &81.88      &88.25\\
SE-ResNet-D (Ours)  &16$\times$16   &72M (68.71G)   &92.27      &\textbf{88.87}\\
\bottomrule
\end{tabular}}
\vspace{-0.3cm}
\end{table}
\begin{figure}[!htb]
  \centering    \centerline{\includegraphics[scale=0.65]{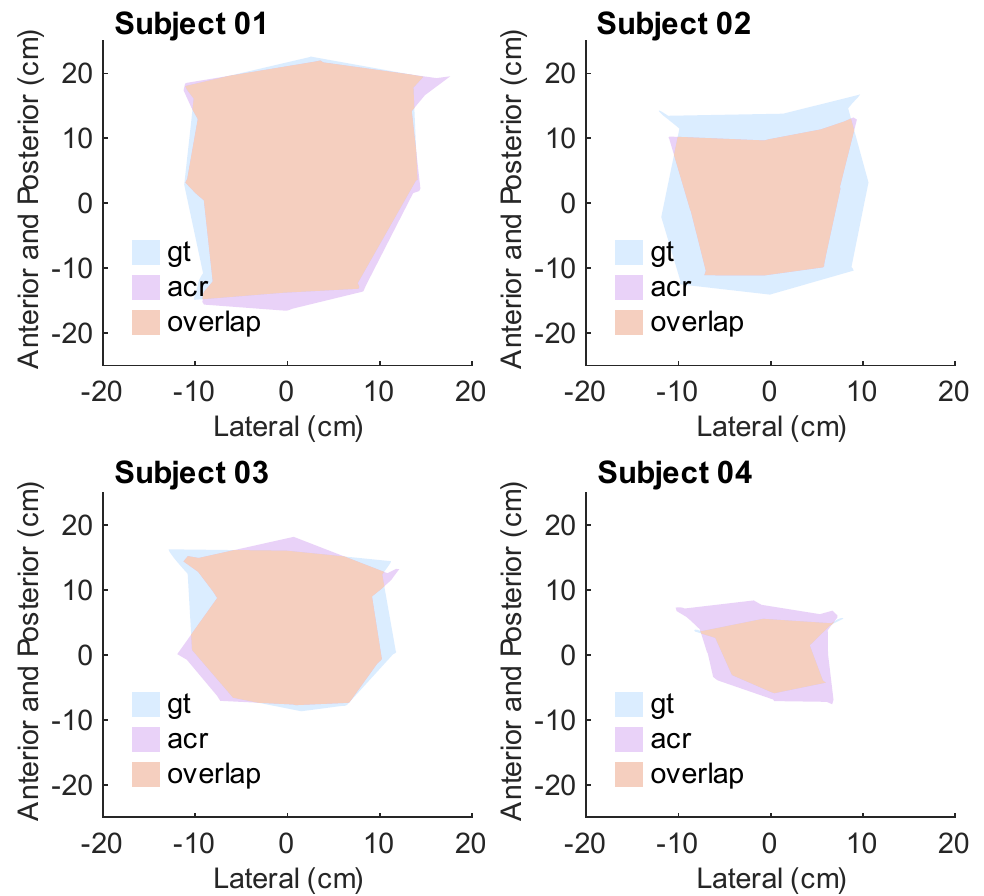}}
    \vspace{-0.1cm}
    \caption{The postural workspace area of 4 subjects performing 1 round movement. $gt$ denotes the ground truth, $acr$ denotes the prediction by our model, and $overlap$ denotes the area covered by both the ground truth and the prediction.}
    \vspace{-0.3cm}
    \label{workspace}
\end{figure}
\begin{figure}[t]  \centering    \centerline{\includegraphics[scale=0.75]{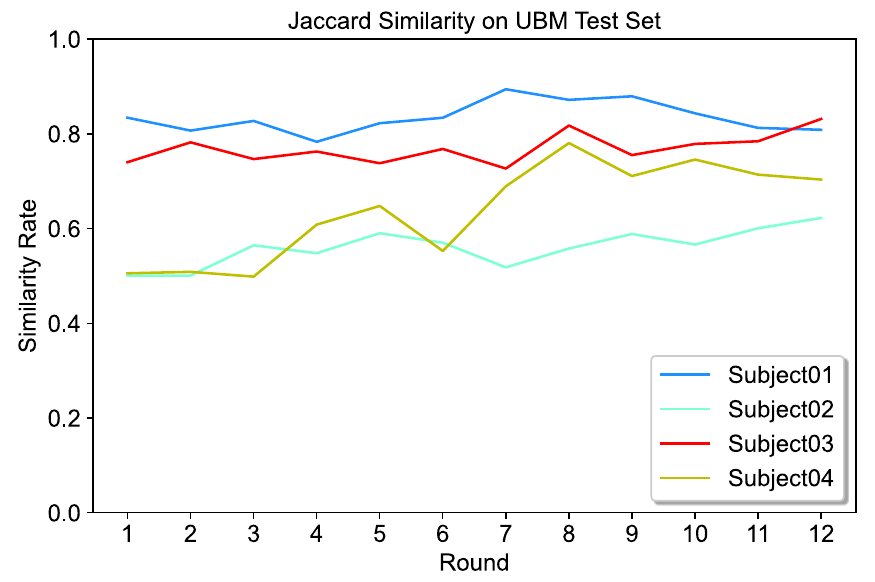}}
 \vspace{-0.2cm}
  \caption{The Jaccard similarity per round of each subject on the UBM test set. This similarity reflects the predictive accuracy of the model. The higher the similarity, the more accurate the model prediction is.}
  \vspace{-0.45cm}
  \label{jaccard}
\end{figure}
\textbf{Backbone Network Architecture:}
We compare several common backbone networks and their variants with ours to validate the effectiveness and efficiency of our backbone. In this experiment, we keep all the other modules in ACRNet the same as what we described in \ref{ACRNetwork}. The results are shown in Table \ref{tab:backbone}. First, we test the performances of ResNet\cite{https://doi.org/10.48550/arxiv.1512.03385} with the different number of layers, and the result shows that the deeper network improves accuracy slightly. Also, the dynamic channel-wise feature attention-based SE-ResNet\cite{hu2018squeeze} is tested. This network promotes accuracy more effectively without adding many parameters than deepening ResNet. We then test the performance from several variants of vanilla SE-ResNet-50 to a backbone designed for ACRNet. In this part, we analyzed the influence of output feature map size, the number of blocks, and activations by expanding the output feature map from $8\times8$ to 16$\times16$ in SE-ResNet-A, changing the number of blocks from (3,4,6,3) to (3,3,9,3) in SE-ResNet-B, substituting GELU activations for ReLU activations in SE-ResNet-C and only reserve the activation between the second and the third layer in each block in SE-ResNet-D (ours) sequentially. The outcome indicates that: 
\begin{itemize}
    \item Expanding output feature map size and increasing block number improve model performance, but the floating point operations (FLOPs), which describe the theoretical, computational complexity of a given mode, also nearly doubled simultaneously.
    \item Substitute GELU activations for ReLU activations lower the mAP a little, but by reducing the number of activations via only reserving the one between the second and the third layer in each block, our backbone reaches its best performance.
\end{itemize}

\subsection{Medical Performance Analysis on our UBM Dataset}
Following the work \cite{santamaria2020robotic}, we apply the postural workspace area as the monitoring index since it intuitively reflects patients' upper body movement ability and postural control limitations. Therefore, we leverage the predicted results by ACRNet of each frame to compute subjects' active sitting workspace area in an actual medical situation, namely the collected UBM dataset, as Fig. \ref{workspace} shows. The horizontal envelope (ignore vertical axis) of 1 round movement is achieved in the following way. First, we follow previous work only using left waist, mid waist, and right waist coordinates to calculate the subject's geometry center by:
\begin{equation}
\left\{
\begin{aligned}
X^i_c &=Mean(X^i_{LW}+X^i_{MW}+X^i_{RW}) \\
Y^i_c &=Mean(Y^i_{LW}+Y^i_{MW}+Y^i_{RW}) 
\end{aligned}
~~~ \forall i \in I
\right.
\end{equation}
where $I$ denotes all frames in that round. Then, the trajectory center $T_c = (\overline{X_c},\overline{Y_c})$  is used to norm the geometry center by:
\begin{equation}
\left\{
\begin{aligned}
X^n_c &=X_c - \overline{X_c} \\
Y^n_c &=Y_c - \overline{Y_c}
\end{aligned}
\right.
\end{equation}
where the final geometry center can be expressed as $G_n = (X^n_c,Y^n_c)$. Finally, we employ Matlab's built-in function $boundary(\cdot)$ to calculate the working area and the boundary.

To evaluate the prediction performance of our model, we apply Jaccard similarity to compare the predicted workspace $P_s$ and the ground truth $GT_s$ as follows:
\begin{align}
\label{Jaccard equaition}
J(P_s,GT_s) = \frac{P_s \cap GT_s}{P_s \cup GT_s}
\end{align}
and the result of each round for all subjects is shown in Fig \ref{jaccard}. It indicates that our model's overall performance is good, whereas the prediction stability of the same subject still needs to be improved. Compared with subjects 1 and 3, the performance is relatively poor on subjects 2 and 4. This situation can be interpreted from 2 aspects. The first is that due to some people's particular movement habits and different body shapes, the similar movement data in the training set is insufficient, which leads to incorrect predictions. Another is that since the belt is manually bound to the subject, there may be errors caused by human factors that make the initial belt position inaccurate, and meanwhile, the belt position might also be shifted during the movement process, resulting in the data collected by Vicon of the same subject vary.

\section{CONCLUSIONS}
This paper proposes a novel and effective Attention Cube Regression network (ACRNet) for 3D human pose estimation from multi-view depth images, which provides a new way to address the low accuracy and high latency problems in telemedicine. By warping the target with an attention cube and calculating the weights of all attention points on each cube surface by the network, the 3D position of each joint is regressed. We elaborate on the structure and the mechanism of each module in ACRNet. To verify our method, we conduct experiments on ITOP, a general HPE dataset, and UBM, a specific medical-related dataset collected by our under real-world rehabilitation environments. A comparison between ACRNet against state-of-the-art methods demonstrates the effectiveness of our method. Extensive experiments justify the design of each module's architecture. We also analyze the performance of ACRNet under the medical monitoring indicator. Besides, experiments show that the high running speed of ACRNet enables it to run in a real-time environment. In the future, we aim to embed this real-time deep learning model into TruST to build a telemedicine platform (e.g., mobile TruST) with accuracy and convenience.

\newcommand{\BIBdecl}{\setlength{\itemsep}{0.12 em}}
\bibliographystyle{IEEEtran}
\bibliography{reference}

\end{document}